%% file: root.tex
\documentclass[letterpaper, 10 pt, conference]{ieeeconf}  

\IEEEoverridecommandlockouts                              

\usepackage{graphics} 
\usepackage{epsfig} 
\usepackage{mathptmx} 
\usepackage{times} 
\usepackage{amsmath} 
\usepackage{amssymb}  
\usepackage{xcolor}
\usepackage{etoolbox}
\usepackage{url}    
\usepackage{fancyhdr}   
\makeatletter
\patchcmd{\@makecaption}
  {\scshape}
  {}
  {}
  {}
\makeatletter
\patchcmd{\@makecaption}
  {\\}
  {.\ }
  {}
  {}
\makeatother

\usepackage{caption}

\usepackage[pagebackref=true,breaklinks=true,letterpaper=true,colorlinks,citecolor=blue,linkcolor=blue,bookmarks=false]{hyperref}

\usepackage{ifthen}
\usepackage{gensymb} 
\usepackage{makecell}
\usepackage{stmaryrd} 
\usepackage{etoolbox}
\makeatletter
\patchcmd{\@makecaption}
  {\scshape}
  {}
  {}
  {}
\makeatletter
\patchcmd{\@makecaption}
  {\\}
  {.\ }
  {}
  {}
\makeatother

\usepackage{caption}
\usepackage{textcomp}

\newboolean{switch_notes}
\setboolean{switch_notes}{false}
\ifthenelse{\boolean{switch_notes}}
{
\newcommand{\note}[2]{\textcolor{green}{\textit{|||[#1: #2]}}}
}{
\newcommand{\note}[2]{}
}

\definecolor{red}{rgb}{0.9,0.1,0}
\definecolor{slateblue}{rgb}{0.7,0.35,0.9}
\definecolor{green}{rgb}{0, 0.4, 0}
\definecolor{orange}{rgb}{1, 0.5, 0}
\definecolor{mahogany}{rgb}{0.75, 0.25, 0.0}
\definecolor{purple}{rgb}{0.6, 0, 0.6}
\definecolor{darkgreen}{rgb}{0, 0.4, 0}
\definecolor{frenchblue}{rgb}{0.0, 0.45, 0.73}
\definecolor{goldenrod}{rgb}{0.85, 0.65, 0.13}
\newboolean{revising}
\setboolean{revising}{false}
\ifthenelse{\boolean{revising}}
{
    \newcommand{\ignore}[1]{}
    
    \newcommand{\albert}[1]{\textcolor{blue}{#1}}
    \newcommand{\kike}[1]{\textcolor{orange}{#1}}

} {
    \newcommand{\ignore}[1]{}
    
    \newcommand{\albert}[1]{#1}
    \newcommand{\kike}[1]{#1}

}

\title{\LARGE \bf
360SD-Net: 360$\degree$ Stereo Depth Estimation with Learnable Cost Volume
}

\author{
Ning-Hsu Wang$^1$, Bolivar Solarte$^1$, Yi-Hsuan Tsai$^3$ \\ Wei-Chen Chiu$^2$, Min Sun$^1$\\
$^1$National Tsing Hua University, $^2$National Chiao Tung University, $^3$NEC Labs America\\
{\tt\small albert100121@gapp.nthu.edu.tw, enrique.solarte.pardo@gmail.com, wasidennis@gmail.com,}
\\
{\tt\small walon@cs.nctu.edu.tw, sunmin@ee.nthu.edu.tw}}

\begin{document}

\maketitle
\thispagestyle{empty}
\pagestyle{empty}

\begin{abstract}
   \albert{Recently, end-to-end trainable deep neural networks have significantly improved stereo depth estimation for perspective images.}
   %
   However, 360$\degree$ images captured under equirectangular projection cannot benefit from directly adopting existing methods due to distortion introduced (i.e., lines in 3D are not projected onto lines in 2D).
   %
  \albert{To tackle this issue, we present a novel architecture specifically designed for spherical disparity using the setting of top-bottom 360$\degree$ camera pairs.}
  Moreover, we propose to mitigate the distortion issue by (1) an additional input branch capturing the position and relation of each pixel in the spherical coordinate, and (2) a cost volume built upon a learnable shifting filter.
    \albert{Due to the lack of 360$\degree$ stereo data, we collect two 360$\degree$ stereo datasets from Matterport3D and Stanford3D for training and evaluation.
    Extensive experiments and ablation study are provided to validate our method against existing algorithms.
    Finally, we show promising results on real-world environments capturing images with two consumer-level cameras.}
    Our project page is at \url{https://albert100121.github.io/360SD-Net-Project-Page}.
\end{abstract}
\input{intro}

\input{related}
\input{method}
\input{exp}
\input{con}

\noindent\textbf{Acknowledgement.}
This project is supported by the National Center for High-performance Computing, MOST
Joint Research Center for AI Technology and All Vista
Healthcare with program MOST 108-2634-F-007-006,
MOST 109-2634-F-007-016 and MOST-109-2636-E-009 -018.
\bibliographystyle{IEEEtran}
\bibliography{root}
\normalsize

\end{document}

%% file: intro.tex
\section{Introduction}
\label{sec:intro}
\begin{figure}
\centering
\includegraphics[width=1\linewidth]{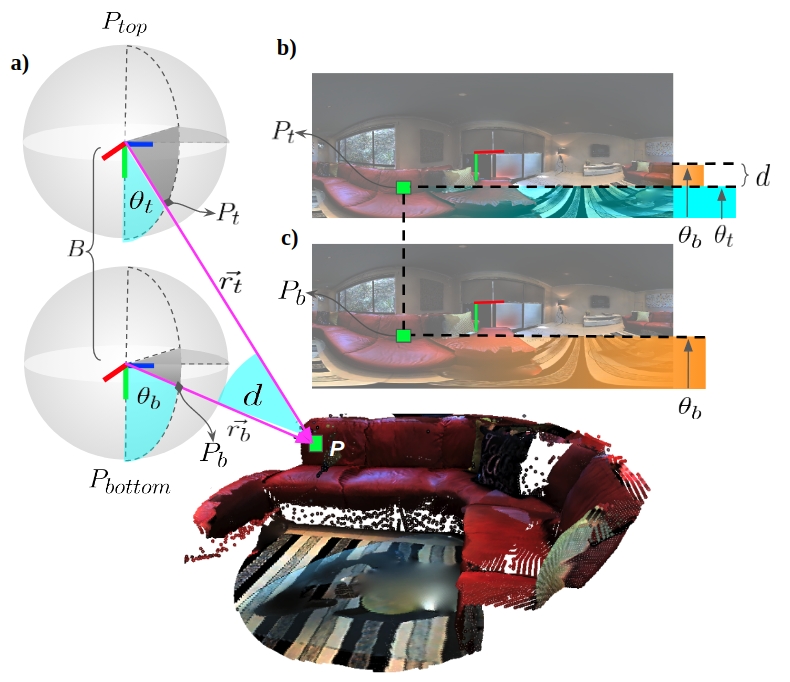}
\caption{Spherical disparity under (a) top-bottom camera pairs ($P_{top}$ and $P_{bottom}$) with baseline $B$. Panel (b)(c) show top and bottom equirectangular projections, respectively. $P_{t}$ and $P_{b}$ are projection points from a 3D point onto the spherical surface (a) and equirectangular coordinate (b)(c). 
$\vec{r}_t$ and $\vec{r}_b$ are projection vectors for the top and bottom cameras, respectively.
$\theta_t$ and $\theta_b$ are the angles between the south pole and its respective projection vector.
$d=\theta_{b}-\theta_{t}$ is the angular disparity. In panel (b)(c), the 3D point projects to the same horizontal position but different vertical positions reflecting the disparity.}
\vspace{-2mm}
\label{fig:sph_disp}
\end{figure}

Stereo depth estimation is a long-lasting yet important task in computer vision due to numerous applications such as autonomous driving, 3D scene understanding, etc.
Despite the majority of studies are for perspective images, disparity can be defined upon various forms of image pairs.
For instance, \kike{the} human binocular disparity is defined as the angle difference between the point of projection on the \albert{retina}, which is part of a sphere rather than a plane.
Similar to human vision, \albert{the} angle difference of a pair of 360$^\circ$ cameras with spherical projection can also be defined as disparity (see Fig.~\ref{fig:sph_disp}(a)).
By taking the advantage of 360$^\circ$ cameras for having a complete observation in an environment, the stereo depth estimation obtained from these cameras enables the 3D reconstruction of \albert{the} entire surrounding.
This is \albert{a} powerful \albert{advantage for advanced applications, e.g., 3D scene understanding.}
%

\setlength{\abovecaptionskip}{0pt}
\setlength{\belowcaptionskip}{0pt}

%
\kike{In this paper, we aim to estimate stereo depth information from a pair of equirectangular images (see Fig.~\ref{fig:sph_disp}(b)(c)), in which they are used in most consumer-level $360^{\circ}$ cameras. For simplicity, we thereafter refer equirectangular images to as $360^{\circ}$ images.
%
}
%
The critical issue needed to cope with is the severe distortion introduced in the process of equirectangular projection. 
First, horizontal lines in 3D are not always the lines in 2D when we use $360^{\circ}$ cameras.
This implies that the typical configuration of the left-right stereo rig may not preserve the \albert{same} property of epipolar \albert{lines}.
%
Therefore, we configure two cameras in a top-bottom manner, such that the epipolar lines on a pair of images are vertically aligned (see Fig. \ref{fig:sph_disp}).
Second, pixels near the top and bottom of the images are stretched more than those located around the equator line. Hence, the corresponding patches at different vertical locations
\albert{are} likely to have different visual characteristics due to different levels of distortion.
This encourages us to propose a novel framework for learning correspondence in top-bottom aligned equirectangular images. 
%
\note{kike}{We could omit "i.e" and the parenthesis... We demonstrate the benefit of each component and compare the performances with respect to the deep-learning baselines PSMNet~\cite{chang2018pyramid} and GCNet \cite{kendall2017end}, additionally, we also compare with conventional stereo matching approaches such as Adaptive Support-Weight (ASW)~\cite{Yoon2006}, Binocular~\cite{li2008binocular}, and PDE-based disparity estimation~\cite{Kim2013}}

We demonstrate the benefit of each component through extensive ablation study and compare the \albert{performance} with deep-learning baselines (i.e., PSMNet~\cite{chang2018pyramid} and GCNet \cite{kendall2017end}) and conventional stereo matching approaches (i.e., 
ASW~\cite{Yoon2006}, Binocular~\cite{li2008binocular}, Kim's~\cite{Kim2013}).
The efficacy of our full model is validated in improving depth estimation for $360^{\circ}$ stereo cameras on two synthetic datasets, as well as generalization to real-world images. The main contributions are as follows:
\begin{itemize}
\item Propose \albert{the first} end-to-end trainable network for stereo depth estimation using $360^{\circ}$ images.
\item Develop a series of \albert{improvements} over existing methods to handle the distortion issue, including the usage of polar angle.
\item Propose a \albert{novel} learnable shifting filter for building \kike{the} cost volume which is empirically better than standard pixel-shifting in the spherical projection.
\item Introduce our $360^{\circ}$ stereo dataset collected from Matterport3D \cite{Matterport3D} and Stanford3D \cite{Stanford3D}, composed of \albert{equirectangular} pairs and depth/disparity ground truths.
\item \albert{Generalize to real-world environments using two consumer-level $360^{\circ}$ cameras with a model trained on the synthetic dataset.}
\end{itemize}

%% file: related.tex
\section{Related Work}
\label{Sec:related}
%
%
%
\subsection{Classical Methods}
Prior to the recent advances of deep learning, numerous research efforts have been devoted to stereo matching and depth estimation. These classical stereo matching algorithms can be roughly categorized into local and global methods.
%
In general, global methods (e.g., Semi-Global Matching (SGM)~\cite{Banz2010}) are able to estimate a better disparity map, but they \albert{need}
to solve a complicated optimization problem. On the other hand, local algorithms (e.g., Adaptive Support-Weight \albert{approach}
(ASW)~\cite{Yoon2006,Hong2017} and Weighted Guided Image Filtering (WGIF)~\cite{Hamzah2018}) are faster and widely used in many embedded applications, but they suffer from the aperture problem or ambiguous matches on homogeneous regions.

\kike{Regarding 360-view methods, Kang \textit{et al.}~\cite{kang19973} target at stereo 360$\degree$ images on a cylinder projection, but do not consider a full 360-view ($4\pi$ steradians).
%
In addition, Im \textit{et al.}~\cite{im2016all} tackle monocular 360$\degree$ depth estimation using structure-from-motion and sphere sweeping algorithm.
Although this model leverages a spherical projection, it is limited to handle short sequences with a high computational cost.
Similar to our setting}, Li \cite{li2008binocular} presents a top-bottom camera setting to define spherical disparity,
while Kim \textit{et al.}~\cite{Kim2013} follow the same \kike{camera} setting but with a PDE-based regularization method to refine the disparity results.
\kike{Although these methods tackle 360$\degree$ stereo depth estimation directly on spherical projection, they still encounter problems of ambiguous matches, artifacts, or diffused surfaces,} where we address them via designing a learning-based framework.

%
\subsection{Deep Learning-based Stereo Method}
Recently, deep learning techniques achieve great progress on stereo \kike{depth estimation}. These techniques can be summarized as a framework with four main components: (1) feature extraction, (2) cost aggregation, (3) cost volume construction, and (4) disparity optimization.
For instance, Koch~\cite{koch2015siamese} and Zbontar \textit{et al.}~\cite{zbontar2016stereo} use a deep metric learning network (e.g., Siamese network) to focus on learning a feature representation in order to obtain better matching cost.
Furthermore, Luo \textit{et al.}~\cite{luo2016efficient} speed up the computation by replacing the concatenation with inner-product for cost aggregation on deep features extracted from stereo pairs.
Considering full-trainable models,  GCNet~\cite{kendall2017end} proposes an end-to-end \kike{deep network}, which has a multi-scale 3D convolution module for producing a more robust disparity regression.
Moreover, PSMNet~\cite{chang2018pyramid} steps further to have spatial pyramid pooling for taking global context information into cost volume and equipping the 3D convolution with a stack of hourglass network to achieve better disparity estimation.
Despite the high performance of the mentioned approaches on stereo perspective views, they do not output desirable results using 360$\degree$ images since properties such as distortion are not considered in their model design.
%
%
\begin{figure*}[t]
\begin{center}
  \centering
\setlength\tabcolsep{2pt}
\begin{tabular}{c}
\makecell{\includegraphics[width=0.9\linewidth]{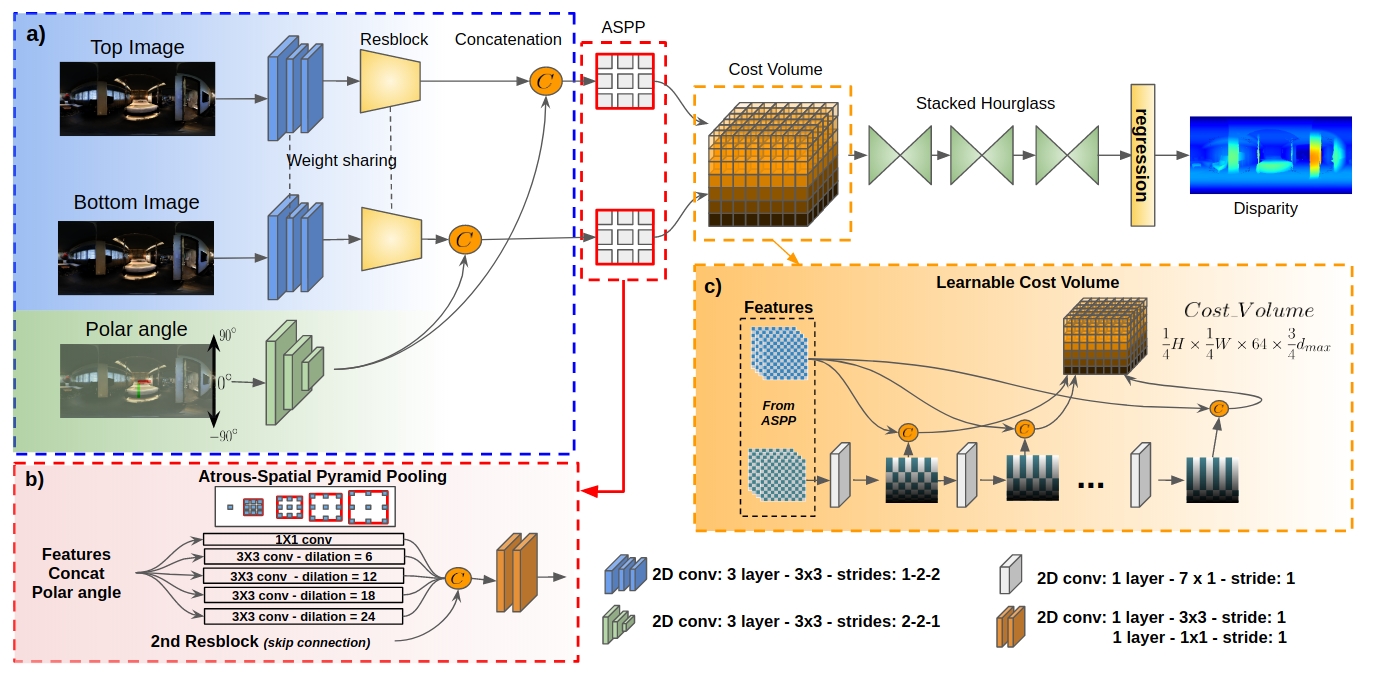}}
\end{tabular}
    \caption{
    Our network mainly consists of three parts: a) two-branch feature extractor that concatenates the stereo equirectangular images and the polar angle in a late fusion setting, b) the ASPP module to enlarge the receptive field, and c) the learnable cost volume to account for the nonlinear spherical projection. Finally, we use the Stacked-Hourglass module to output the final disparity map.}
    \label{fig:archi}
\end{center}
\end{figure*}
\subsection{Vision Techniques for 360\degree Camera}
When the consumer-level $360^{\circ}$ cameras were made easily available and affordable, it attracts significant research interest from the computer vision and robotics communities.
\albert{For instance, Cohen \textit{et al.}~\cite{DBLP:journals/corr/abs-1801-10130} and Esteves \textit{et al.}~\cite{esteves2018learning} process spherical information on spectral-domain for classification, whereas KTN~\cite{DBLP:journals/corr/abs-1812-03115} and Flat2sphere~\cite{DBLP:journals/corr/abs-1708-00919} focus on designing spherical convolution kernels such that the network can support multiple recognition tasks in $360^{\circ}$ images.}
\kike{On the other hand, several works~\cite{zou2018layoutnet, yang2019dula, Sun_2019_CVPR} leverage 360-views to reconstruct layout scenes from equirectangular images as input. 
Similarly, \cite{Zhang_2018_ECCV, Cheng_2018_CVPR} address the problem of saliency detection in $360^{\circ}$ videos for exploring the rich content of a scene in a more efficient manner using the full view of equirectangular representation and dealing with distortion properly. 
For depth estimation purposes, \cite{Zioulis_2018_ECCV, DBLP:journals/corr/abs-1811-05304, Garanderie_2018_ECCV} tackle monocular depth estimation from $360^{\circ}$ images via leveraging re-projection models, rendered scenes, and structures-from-motion techniques.}
%
\albert{Recently, SweepNet~\cite{sweepnet} targets multi-view stereo depth \kike{estimation applying} a deep network on four fish-eye images re-projected into \kike{concentric virtual spheres} to estimate 360$\degree$ depths.}

\albert{
\kike{Despite the previous approaches,} there exists a literature gap in 360$\degree$ stereo-depth estimation using convolutional networks.
Therefore, we provide a novel deep network, 
\kike{which relies on two equirectangular images as input and deals with distortion effectively.}
\albert{Such input is the minimum requirement for a stereo setup that keeps the benefits of a full 360$\degree$ view~\cite{Kim2013, li2008binocular}. \kike{Moreover, our proposed model is capable of being applied directly by commercial-level 360$\degree$ cameras, making this solution highly affordable.} }
To the best of our knowledge, we are the first to target at deep learning-based stereo depth estimation from 360$\degree$ images.} 
We note~\cite{Eder_2019,Zioulis_2019} as concurrent works with similar idea to our paper.
\vspace{-1mm}

%% file: method.tex
\section{Method}
\label{section:Method}
%
%
The proposed framework, namely 360SD-Net, 
investigates a unique stereo depth estimation pipeline for 360$\degree$ cameras.
We first introduce our camera setting and define the spherical disparity.
Then, we
\albert{propose} the end-to-end trainable model as depicted in Fig.~\ref{fig:archi}.
\note{kike}{another way to introduces fig 2 would be better}
%

\subsection{Camera Setting and Spherical Disparity}
\label{subsection:camera_setting}
We use a top-bottom camera setting (similar to~\cite{li2008binocular,Kim2013}), where the stereo correspondence lies on the same vertical line on the camera spheres (see Fig.~\ref{fig:sph_disp}(a)).
This setting also ensures that the correspondence lies on the same vertical line in 360$\degree$ images captured under equirectangular projection (see Fig.~\ref{fig:sph_disp}(b,c)). Our setting can be built with relatively low cost since most consumer-level 360$\degree$ cameras capture images under equirectangular projection.

\albert{We now define spherical disparity using the following terms (see Fig.~\ref{fig:sph_disp}). $P_{t}$ and $P_{b}$ are projection points from a 3D point $P$ onto the camera sphere of the top and bottom camera, respectively.
$\vec{r}_t$ and $\vec{r}_b$ are projection vectors, while $\theta_t$/$\theta_b$ are the angles between the south pole and $\vec{r}_t$/$\vec{r}_b$ for the top and bottom cameras, respectively. The disparity is defined as the difference between the two angles with following equation $d=\theta_{b}-\theta_{t}$.}
The depth with respect to the top camera equals to the norm of $\vec{r}_t$\albert{, which} is computed as follows,
\begin{equation}
\left |\Vec{{r}_t}  \right | = 
B \cdot \left [ \frac{\sin{(\theta_{t}})}{\tan{(d)}} + \cos{(\theta_{t})} \right ]~,
\label{eq:spherical_disparity}
\end{equation}
where $B$ is the baseline between top and bottom cameras.
%
%
%
Note that the disparity and depth relation is not fixed as in perspective stereo cameras, but varies according to the angle $\theta_t$. Hence, the meaning of disparity estimation error becomes less intuitive. In practice, we mainly evaluate depth instead of disparity estimation. 
%

\subsection{Incorporation with Polar Angle}
As described \albert{in Section~\ref{Sec:related}, deep stereo depth estimation}
\albert{methods~\cite{chang2018pyramid, zbontar2016stereo, kendall2017end}} disregard distortion introduced in equirectangular images. \albert{To address this problem, we add the polar angle (see Fig.~\ref{fig:archi}(a)) as the model input for additional geometry information since it is closely related to the distortion.}
%
\albert{
In order to separate geometry information from the RGB appearance information, we apply residual blocks for RGB input and three Conv2D layers for polar angle instead of directly concatenating model input (i.e., early fusion design).
%
%
Then, both outputs are concatenated after feature extraction, in which we refer to this procedure as our late fusion design.
The comparison of both designs is shown in the experimental section.}
%
%
\subsection{ASPP Module}
After \albert{fusing} image features with the geometry information, we still have to \albert{manage} the spatial relationship among pixels, \albert{since} $360^{\circ}$ images provide a larger field-of-view than 
\albert{regular} images.
%
\albert{In order to consider} different scales spatially, we adopt recent advances ASPP~\cite{chen2014semantic} as proposed for semantic segmentation (see Fig.~\ref{fig:archi}(b)). This module is a dilated convolution design considering multi-scale resolutions at different levels of the receptive field.
In order to reduce the large memory consumption for cost volume-based stereo depth estimation, we perform random cropping during training. 
%
%
\subsection{Learnable Cost Volume}
\albert{The following critical step for stereo matching is to construct a 3D cost volume by computing the matching costs at a pre-defined disparity levels with a fixed step-size.
This step-size in a typical 3D cost volume is one pixel, i.e., approaches like GCNet \cite{kendall2017end} and PSMNet \cite{chang2018pyramid} concatenate left and right features to construct 3D cost volume based on one-pixel step-size. However, with the distortion introduced by equirectangular projection, per-pixel step-size is not consistent with the geometry information from the polar angle input. Under this premise, we introduce a novel learnable cost volume (LCV) in our 360SD-Net using a shifting filter, which searches the optimal step-size on ``degree unit'' in order to precisely construct the optimal cost volume.
}
%

\albert{We design our LCV with a shifting filter via a $7 \times 1$ Conv2D layer, as shown in Fig.~\ref{fig:archi}(c), and apply channel-wise shifting with the proposed filter to prevent the mixture between channels.
This filter design allows vertical shifting to satisfy our stereo setting and retains the full view of the equirectangular images.
Therefore, the best shifting step-size of the feature map would be learned by convolution.
Note that, we apply replicated-padding instead of zero-padding before each convolution to retain the boundary information.
To ensure stable training in practice, we still follow the normal cost volume shifting (freezing the parameters for the shifting Conv2D) in the first 50 epochs and start learning the cost volume shifting afterward.}

%
%
\subsection{\albert{3D Encoder-Decoder and }Regression Loss}
\albert{We adopt the stacked hourglass~\cite{chang2018pyramid} as our 3D Encoder-Decoder and the regression as in 
\cite{kendall2017end} to regress continuous disparity values.
It is reported that this disparity regression is more robust than classification-based stereo depth estimation methods.
For the loss function, we 
use the smooth L1 loss with the ground truth disparity.}

%% file: exp.tex
\section{Experimental Results}

\begin{table*}[t]
\caption{\small{Experimental results of the proposed method on MP3D and SF3D compared with other approaches including deep learning-based networks and conventional algorithms.($\shortdownarrow$ represents the lower the better.)}}
\begin{center}
\begin{tabular}{ l || l l l l l| l l l l l l }
\hline
\multicolumn{1}{c||}{} & 
\multicolumn{5}{c|}{\textbf{MP3D}} & 
\multicolumn{5}{c}{\textbf{SF3D}} \\
\multicolumn{1}{c||}{} &
\multicolumn{2}{c}{\textbf{Disparity}}     & 
\multicolumn{2}{c}{\textbf{Depth}} &
\multicolumn{1}{c|}{\textbf{Time}} &
\multicolumn{2}{c}{\textbf{Disparity}}     & 
\multicolumn{2}{c}{\textbf{Depth}} &
\multicolumn{1}{c}{\textbf{Time}} \\
\multicolumn{1}{c||}{}& \textbf{MAE}~$\shortdownarrow$ & \textbf{RMSE}~$\shortdownarrow$ 
& \textbf{MAE}~$\shortdownarrow$ & \textbf{RMSE}~$\shortdownarrow$ & &
\textbf{MAE}~$\shortdownarrow$ &  \textbf{RMSE}~$\shortdownarrow$ 
& \textbf{MAE}~$\shortdownarrow$ & \textbf{RMSE}~$\shortdownarrow$ &\\ \hline \hline
Binocular \cite{li2008binocular}    &0.7206  &2.507
&0.1368 &0.5399 &0.6333 
&0.3204  &1.5494  &0.0897 &0.4496 &0.6333 \\ \hline
KIM's \cite{Kim2013}    &0.8175  &2.2956  
&0.2191 &0.6955 &1.8507 
&2.5327  &4.39  &0.1163 &0.3972 &1.8507 \\ \hline
ASW \cite{Yoon2006}             &0.4410  &1.648 &0.1427 &0.5193 &7.5min 
&0.2155  &0.7754  &0.0779 &0.2628 &7.5 min \\ \hline
GCNet \cite{kendall2017end}                  &0.486  &1.4283  
&0.0969  &0.2953 & 1.54s 
&0.1877  &0.4971  &0.0592  &0.1361 &1.57s \\ \hline
PSMNet \cite{chang2018pyramid}                  &0.3139  &1.049  
&0.0946  &0.2838 & \textbf{0.50s}
&0.1292  &0.4053  &0.0418  &0.1068 &\textbf{0.51s}  \\ \hline
360SD-Net (Ours) 
&\textbf{0.1447}  &\textbf{0.6930} 
&\textbf{0.0593}   &\textbf{0.2182}  &0.572s %
&\textbf{0.1034}  &\textbf{0.3691} &\textbf{0.0335}   &\textbf{0.0914} & 0.55s\\ \hline
\end{tabular}
\label{table:MP3D_SF3D_baseline}
\end{center}
\end{table*}
\subsection{\albert{Dataset and System Configuration}}
\albert{Due to the lack of 360$\degree$ stereo dataset, we have collected two photo-realistic datasets MP3D and SF3D
through Matterport3D~\cite{Matterport3D} with Minos virtual environment~\cite{savva2017minos} and re-projection of Stanford3D point clouds~\cite{Stanford3D}.}
%
\albert{Considering the complexity and the extensive effort required to stitch, calibrate, and collect real-world RGB images and depth maps, which is not suitable for the training of deep models, we train our model solely on the presented synthetic data.}
%

\albert{The setting of our dataset is a pair of 360$\degree$ top-bottom aligned stereo images with equirectangular projection. The resolution of these images is 512 in height and 1024 in width, which is commonly used in 360$\degree$ works~\cite{Sun_2019_CVPR,Cheng_2018_CVPR,wang2018self}.}
\albert{The baseline of our stereo system is set to 20 cm, and the number of data we have collected in MP3D/SF3D datasets for training, validation, and testing are 1602/800, 431/200, 341/203, respectively. Each data consists of four components, a RGB-image pair, depth, and disparity. For data collection, we have diversified indoor scenarios in each set to prevent similarities and repetitiveness.}
%
Furthermore, the two datasets and code will be made available to the public.
%
\begin{figure}[]
  \centering
\setlength\tabcolsep{2pt}
\begin{tabular}{c}
\makecell{\includegraphics[width=0.98\linewidth]{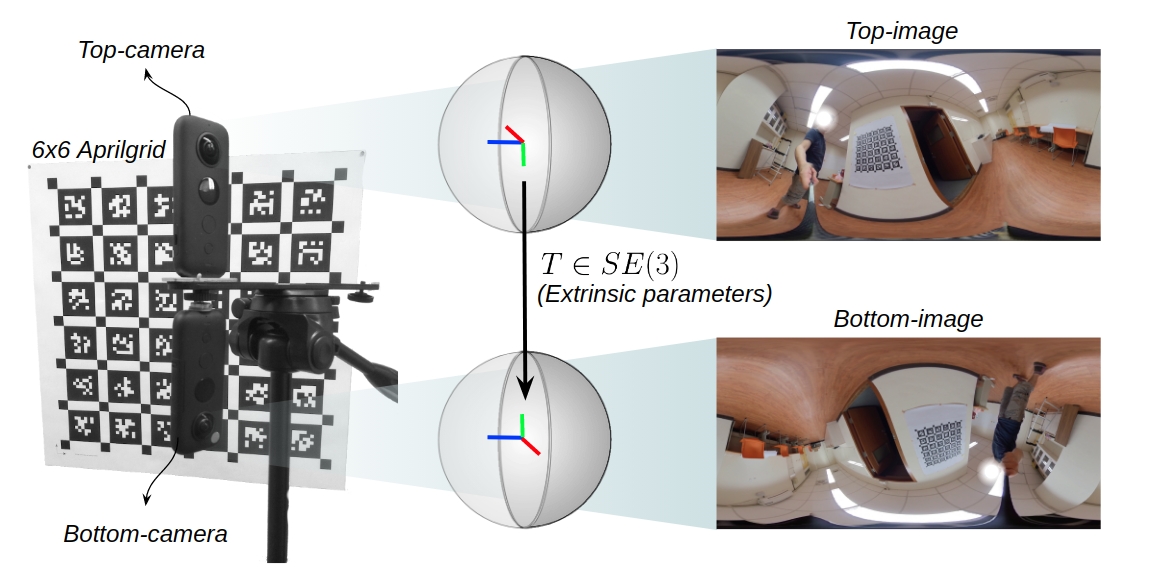}}
\end{tabular}
    \caption{\kike{Our 360$\degree$ stereo system composed of two  Insta360\textsuperscript{\textregistered} ONE X cameras. In order to align both equirectangular images (top and bottom), the extrinsic parameters between the cameras is needed. This transformation is obtained by stereo calibration.}}
    \label{fig:Stereo_camera_system}
\end{figure}

\albert{We also provide qualitative results on real-world scenes to show the generalization of our model between synthetic training and real-world testing.
These real-world scenes are collected with two well-known consumer-level 360$\degree$ cameras, Insta360\textsuperscript{\textregistered} ONE X (Fig.~\ref{fig:Stereo_camera_system}).
Both cameras are calibrated using a 6x6 Aprilgrid and the toolbox calibration \textit{Kalibr}, in particular~\cite{kannala2006generic}. On the other hand, to preserve our camera setting described in Section~\ref{subsection:camera_setting}, we align the polar axis of both equirectangular images using the extrinsic transform obtained by the calibration.}

\subsection{Metrics} \label{metric_sec}
\albert{We have evaluated both depth and disparity results using MAE and RMSE.
The depth error is prone to increasing significantly based on the non-linear relationship between depth and disparity as in \eqref{eq:spherical_disparity}, which does not provide informative evaluation. Therefore, we crop out 5\% of largely distorted depth map from the top and bottom, respectively.}
%
\subsection{\albert{Experimental Setting}}
\label{exper_settings}
\albert{Our model is trained from scratch with Adam ($\beta_1=0.9,\beta_2=0.999$) solver for 400 epochs with an initial learning rate of 0.001 and fine-tuned with a
learning rate of 0.0001 for 100 epochs on MP3D. For SF3D, we follow the same setting as MP3D but with 50 epochs using pre-trained model from MP3D.}
The entire implementation is based on the PyTorch framework.

%
\begin{table}[]
\centering
\caption{\small{Ablation study for depth estimation on MP3D. The first row \textbf{bs} is considered as the baseline in this study\albert{, which uses a fixed step-size vertical pixel shifting.}. Different components are denoted as: (\textbf{Pc}) Polar angle with early fusion; (\textbf{Pb}) Polar angle with late fusion; (\textbf{ASPP}) ASPP module; (\textbf{LCV}) Learnable Cost Volume; (\textbf{repli}) LCV with replicate padding.}}
\begin{tabular}{lllc}
\hline
\textbf{ID} & \textbf{Ablation Study} & \textbf{Depth RMSE}~$\shortdownarrow$  \\ \hline
1&bs & 0.2765 \\ \cline{2-3}
2&bs + Pc (Table~\ref{table:polar_LCV_ablation} Coordinate ID2)& 0.2501  \\ \cline{2-3}
3&bs + Pb & 0.2494 (+9.8\%) \\ \cline{2-3}
4&bs + Pb + ASPP & 0.2462 (+10.9\%) \\ \cline{2-3}
5&LCV  (Table~\ref{table:polar_LCV_ablation} Step Size ID3)& 0.2464\\ \cline{2-3}
6&LCV (repli) & 0.2409 (+12.9\%)\\ \cline{2-3}
7&(Ours) bs + Pb + ASPP + LCV (repli) & \textbf{0.2182} (+21.1\%) \\ \cline{2-3}
\end{tabular}
\label{table:exp_ablation}
\end{table}
\begin{table}[]
\centering
\scriptsize
\caption{\small{Ablation study of different coordinate information added and different initial step-size of LCV for depth estimation on MP3D.}}
\begin{tabular}{lll || ll}
\hline
\textbf{ID} & \textbf{Coordinate} & \textbf{Depth RMSE}~$\shortdownarrow$ & \textbf{Step Size} & \textbf{Depth RMSE}~$\shortdownarrow$ \\ \hline
1&horizontal angle  & 0.2583 & 1$\degree$     & 0.2611 \\ \cline{2-5}
2&polar angle   & \textbf{0.2501} &
1/2~$\degree$& 0.2559  \\ \cline{2-5}
3&radius    & 0.2541 &
1/3~$\degree$ &\textbf{0.2464}  \\ \cline{2-5}
4&arc-length & 0.2516 &
1/4~$\degree$& 0.2503 \\ \cline{2-5}
5&area      & 0.2513 & - & -\\ \cline{2-5}
\cline{2-5}
\end{tabular}
\label{table:polar_LCV_ablation}
\end{table}
\subsection{Overall Performance}
%
\albert{In Table~\ref{table:MP3D_SF3D_baseline}}, we show results on MP3D and SF3D with comparisons to state-of-the-art stereo
\albert{depth estimation} approaches, including the conventional methods (ASW~\cite{Yoon2006}, Binocular~\cite{li2008binocular} and KIM's~\cite{Kim2013}) and deep learning-based models (PSMNet~\cite{chang2018pyramid} and GCNet~\cite{kendall2017end}).
Our method achieves significant improvement for both the disparity and depth performances\albert{, since other methods do not consider distortion introduced in 360$\degree$ images}.
These results demonstrate the effectiveness of our designs for 360$\degree$ images\albert{, including polar angle and LCV modules}.
In addition, compared to the baseline PSMNet model, our method only introduces a slight overhead in runtime, while our model outperforms PSMNet by a large margin.
    
\begin{figure}[]
  \centering
\setlength\tabcolsep{2pt}
\begin{tabular}{c}
\makecell{\includegraphics[width=0.98\linewidth]{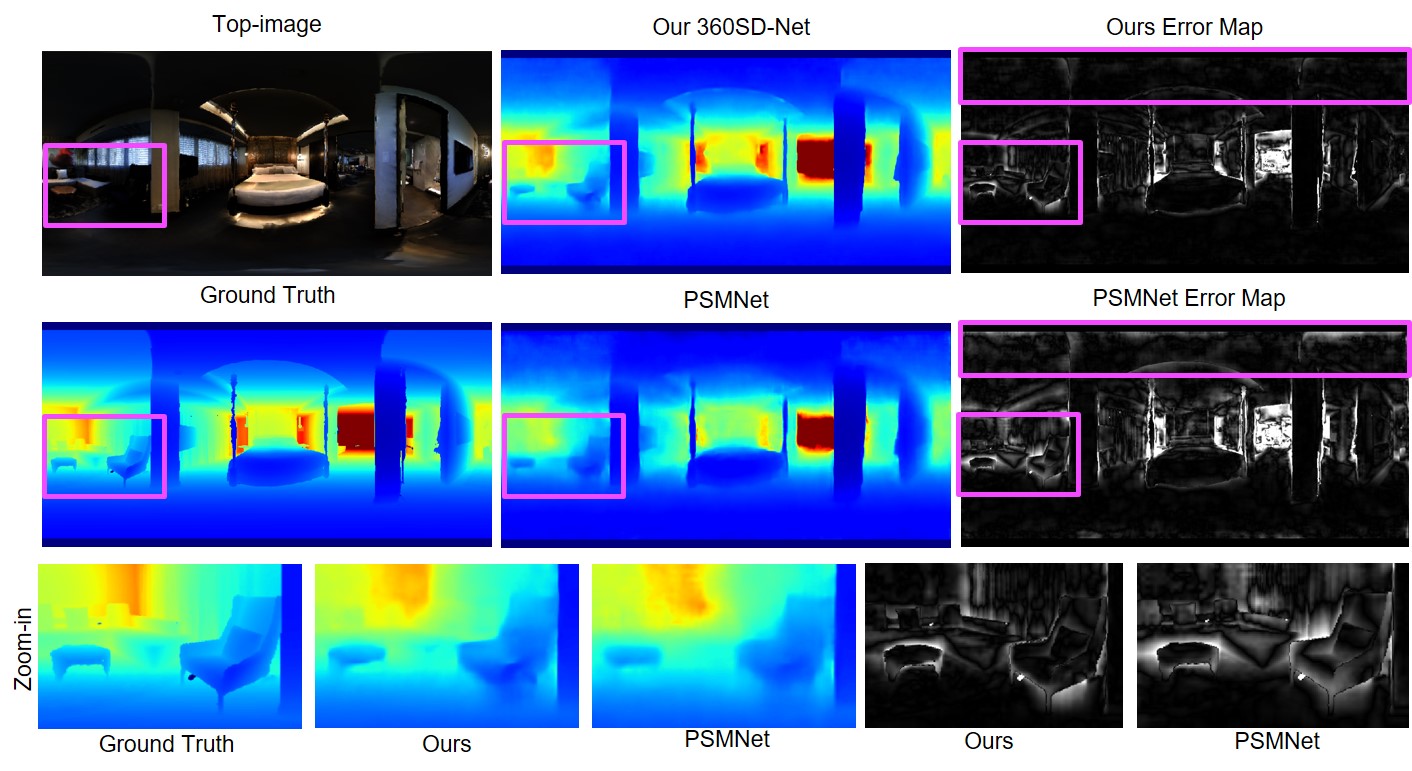}}
\end{tabular}
    \caption{\kike{Qualitative depth map and error map comparison between 360SD-Net (Ours) and PSMNet.} \albert{Our depth map shows sharper and clearer details in both close and distant regions. For the \kike{zoom-in views}, the armchair and table \kike{present a notable geometry structure compared to the one from PSMNet. Our error map also shows higher accuracy in regions with higher distortion and object boundaries.}}}
    
    \label{fig:Depth_PCL}
\end{figure}
%
\begin{figure}[]
  \centering
\setlength\tabcolsep{2pt}
\begin{tabular}{c}
\makecell{\includegraphics[width=0.98\linewidth]{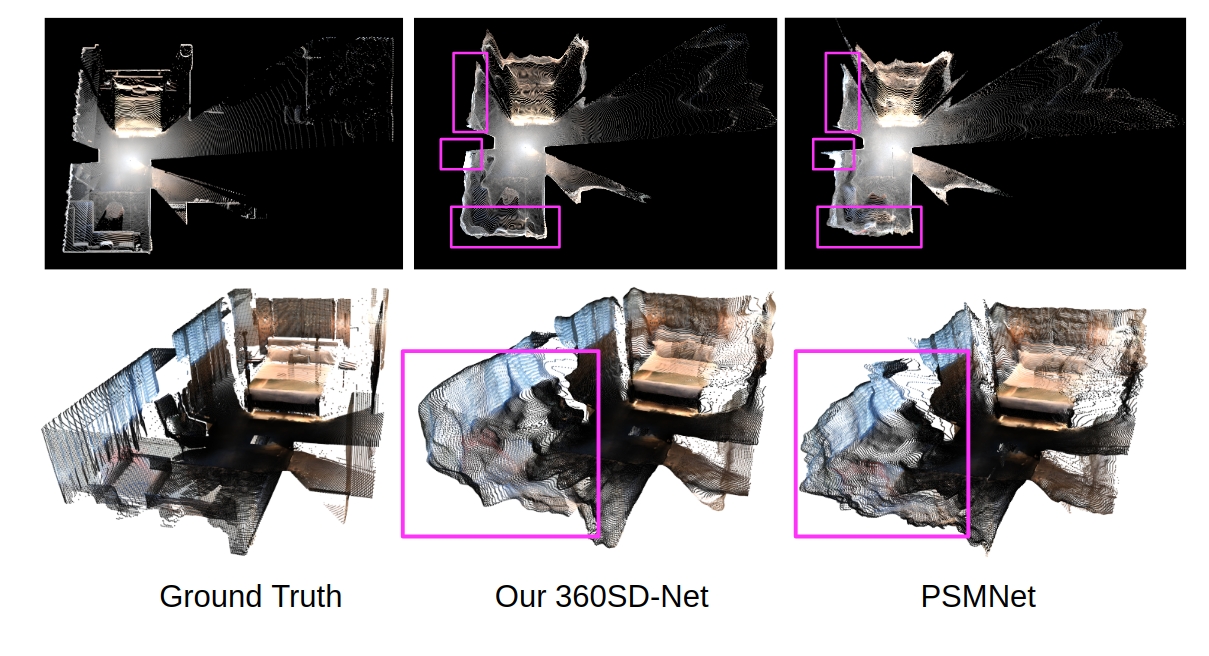}}
\end{tabular}
    \caption{Qualitative point cloud comparison between 360SD-Net (Ours) and the PSMNet. Our model shows a better geometry estimation with less distortion and a more accurate structure.}
    \label{fig:PCL}
\end{figure}
\subsection{Ablation Study} \label{Ablation study}
We present an ablation study in Table~\ref{table:exp_ablation} on MP3D for depth estimation to validate the effectiveness of each component in the proposed framework. \albert{Comparing ID 1 with 5, it shows the effectiveness of LCV, while ID 2 shows the benefits from the polar angle. 
The other rows gradually show the improvement of adding other designs such as ASPP and replicated-padding. With the combination of ID 4 and 6, we form our final network that achieves the best performance.}

\albert{Detailed ablation studies on polar angle and LCV are shown in Table~\ref{table:polar_LCV_ablation},
which compares different geometry measurements from spherical projection and different initial step-sizes in degree applied in LCV.
Through comparing various geometry measurements, including area, arc-length, radius, and horizontal angle, using the polar angle performs the best in dealing with distortion.
%
Regarding initial step-size in LCV, we demonstrate empirically that the performance increases when the initial step-size value decreases.
The improvements saturate at $\frac{1}{3}\degree$, which is chosen to be our initial step-size value.}
\begin{figure*}[]
   \centering
\setlength\tabcolsep{2pt}
\begin{tabular}{c}
\makecell{\includegraphics[width=0.9\linewidth]{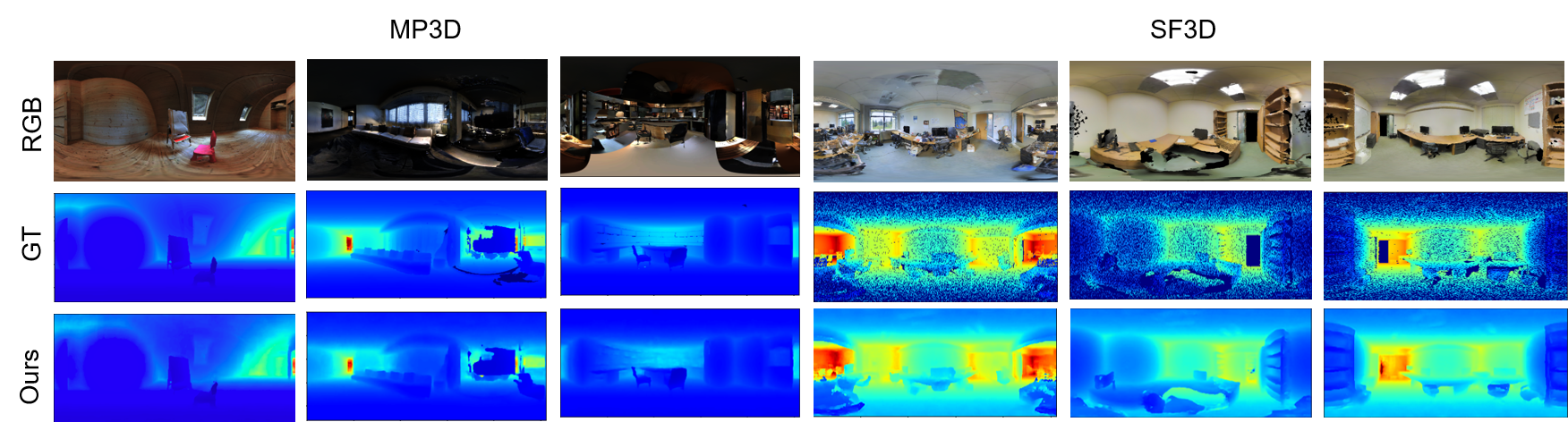}}
\end{tabular}
    \caption{More qualitative results for depth map on MP3D and SF3D. For MP3D, our estimated depth maps preserve object and surface details with results similar to GT. For SF3D, our model outputs dense depth maps of high accuracy, with training on sparse GT.}
    \label{fig:More_depth_MP3D_SF3D}
\end{figure*}
\begin{figure*}[]
  \centering
\setlength\tabcolsep{2pt}
\begin{tabular}{c}
\makecell{\includegraphics[width=0.9\linewidth]{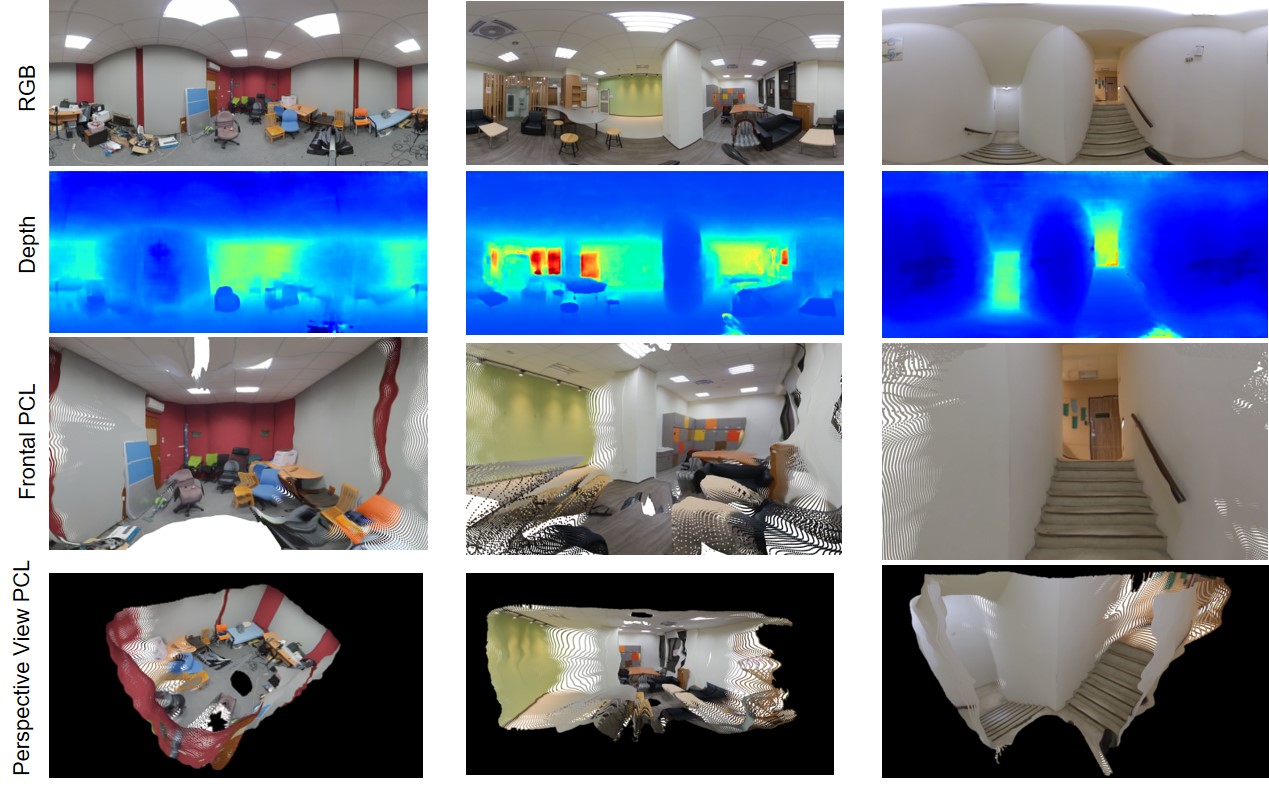}}
\end{tabular}
    \caption{
    Qualitative results on real scenes using two Insta360\textsuperscript{\textregistered} ONE X cameras in a top-bottom configuration. The furniture can be clearly seen in the depth maps and also well reconstructed in the point clouds.}
    \label{fig:Real_pcl}
\end{figure*}
%
%
%
%
\subsection{Qualitative Results}
\albert{We present qualitative results in depth maps and point clouds, mainly compared with PSMNet~\cite{chang2018pyramid} based on its good performance in \albert{Table~\ref{table:MP3D_SF3D_baseline}}. As shown in \albert{Fig.~\ref{fig:Depth_PCL} and Fig.~\ref{fig:PCL}},
our model results in sharper depth maps and our projected point clouds are able to reconstruct scenes more accurately in comparison to PSMNet.
\albert{Furthermore, Fig.~\ref{fig:More_depth_MP3D_SF3D}} shows more qualitative results 
on both datasets of our model.}
%
%
\subsection{Qualitative Results for Real-World Images}
\albert{To show the generalization of our model (trained on the synthetic MP3D dataset) on real-world scenes, we take \albert{still and moving images}
with a pair of well-known consumer-level 360$\degree$ cameras.
In order to reduce the domain gap, we apply our model on these real-world images using gray-scale.
%
%
%
In Fig.~\ref{fig:Real_pcl}, we show 
the results in depth maps, frontal view, and perspective view of point clouds with their regarded RGB images.
The details of objects and room layouts are elegantly reconstructed, which shows great compatibility of our network between synthetic and real-world scenes.
%
%
Moreover, our model produces promising depth maps for handheld videos (refer to supplementary video for more results).}
%

%% file: con.tex
\section{Conclusions}
\albert{In this paper, we introduce the first end-to-end trainable deep network, namely 360SD-Net, for depth estimation directly on 360$\degree$ stereo images via designing a series of improvements over existing methods.
In experiments, we show state-of-the-art performance on our collected synthetic datasets with extensive ablation study that validates proposed modules, including the usage of polar angle and learnable cost volume design. Finally, we test on real-world scenes and present promising results with the model trained on the pure synthetic data to show the 
\albert{generalization} and {compatibility} of our presented network.}
%
%

%% file: root.bbl
\begin{thebibliography}{10}
\providecommand{\url}[1]{#1}
\csname url@rmstyle\endcsname
\providecommand{\newblock}{\relax}
\providecommand{\bibinfo}[2]{#2}
\providecommand\BIBentrySTDinterwordspacing{\spaceskip=0pt\relax}
\providecommand\BIBentryALTinterwordstretchfactor{4}
\providecommand\BIBentryALTinterwordspacing{\spaceskip=\fontdimen2\font plus
\BIBentryALTinterwordstretchfactor\fontdimen3\font minus
  \fontdimen4\font\relax}
\providecommand\BIBforeignlanguage[2]{{%
\expandafter\ifx\csname l@#1\endcsname\relax
\typeout{** WARNING: IEEEtran.bst: No hyphenation pattern has been}%
\typeout{** loaded for the language `#1'. Using the pattern for}%
\typeout{** the default language instead.}%
\else
\language=\csname l@#1\endcsname
\fi
#2}}

\bibitem{chang2018pyramid}
J.-R. Chang and Y.-S. Chen, ``Pyramid stereo matching network,'' in
  \emph{Proceedings of the IEEE Conference on Computer Vision and Pattern
  Recognition}, 2018, pp. 5410--5418.

\bibitem{kendall2017end}
A.~Kendall, H.~Martirosyan, S.~Dasgupta, P.~Henry, R.~Kennedy, A.~Bachrach, and
  A.~Bry, ``End-to-end learning of geometry and context for deep stereo
  regression,'' in \emph{Proceedings of the International Conference on
  Computer Vision ({ICCV})}, 2017.

\bibitem{Yoon2006}
K.~J. Yoon and I.~S. Kweon, ``{Adaptive support-weight approach for
  correspondence search},'' \emph{IEEE Transactions on Pattern Analysis and
  Machine Intelligence}, vol.~28, 2006.

\bibitem{li2008binocular}
S.~Li, ``Binocular spherical stereo,'' \emph{IEEE Transactions on intelligent
  transportation systems}, vol.~9, no.~4, pp. 589--600, 2008.

\bibitem{Kim2013}
H.~Kim and A.~Hilton, ``3d scene reconstruction from multiple spherical stereo
  pairs,'' \emph{International Journal of Computer Vision}, vol. 104, 08 2013.

\bibitem{Matterport3D}
A.~Chang, A.~Dai, T.~Funkhouser, M.~Halber, M.~Niessner, M.~Savva, S.~Song,
  A.~Zeng, and Y.~Zhang, ``{Matterport3D}: Learning from {RGB-D} data in indoor
  environments,'' \emph{International Conference on 3D Vision (3DV)}, 2017.

\bibitem{Stanford3D}
I.~{Armeni}, A.~{Sax}, A.~R. {Zamir}, and S.~{Savarese}, ``{Joint
  2D-3D-Semantic Data for Indoor Scene Understanding},'' \emph{ArXiv e-prints},
  Feb. 2017.

\bibitem{Banz2010}
C.~Banz, S.~Hesselbarth, H.~Flatt, H.~Blume, and P.~Pirsch, ``{Real-time stereo
  vision system using semi-global matching disparity estimation: Architecture
  and FPGA-implementation},'' in \emph{2010 International Conference on
  Embedded Computer Systems: Architectures, Modeling and Simulation}, 2010.

\bibitem{Hong2017}
G.~S. Hong and B.~G. Kim, ``{A local stereo matching algorithm based on
  weighted guided image filtering for improving the generation of depth range
  images},'' \emph{Displays}, 2017.

\bibitem{Hamzah2018}
\BIBentryALTinterwordspacing
R.~A. Hamzah, M.~S. Hamid, A.~F. Kadmin, S.~F.~A. Gani, S.~Salam, and T.~M.
  Wook, ``{Accurate Disparity Map Estimation Based on Edge-preserving
  Filter},'' in \emph{2018 International Conference on Smart Computing and
  Electronic Enterprise, ICSCEE 2018}.\hskip 1em plus 0.5em minus 0.4em\relax
  IEEE, 2018. [Online]. Available:
  \url{https://ieeexplore.ieee.org/document/8538360/}
\BIBentrySTDinterwordspacing

\bibitem{kang19973}
S.~B. Kang and R.~Szeliski, ``3-d scene data recovery using omnidirectional
  multibaseline stereo,'' \emph{International journal of computer vision},
  vol.~25, no.~2, pp. 167--183, 1997.

\bibitem{im2016all}
S.~Im, H.~Ha, F.~Rameau, H.-G. Jeon, G.~Choe, and I.~S. Kweon, ``All-around
  depth from small motion with a spherical panoramic camera,'' in
  \emph{European Conference on Computer Vision}.\hskip 1em plus 0.5em minus
  0.4em\relax Springer, 2016, pp. 156--172.

\bibitem{koch2015siamese}
G.~Koch, ``Siamese neural networks for one-shot image recognition,'' 2015.

\bibitem{zbontar2016stereo}
J.~Zbontar and Y.~LeCun, ``Stereo matching by training a convolutional neural
  network to compare image patches,'' \emph{Journal of Machine Learning
  Research}, vol.~17, pp. 1--32, 2016.

\bibitem{luo2016efficient}
W.~Luo, A.~G. Schwing, and R.~Urtasun, ``Efficient deep learning for stereo
  matching,'' in \emph{Proceedings of the IEEE Conference on Computer Vision
  and Pattern Recognition}, 2016, pp. 5695--5703.

\bibitem{DBLP:journals/corr/abs-1801-10130}
\BIBentryALTinterwordspacing
T.~S. Cohen, M.~Geiger, J.~K{\"{o}}hler, and M.~Welling, ``Spherical cnns,''
  \emph{CoRR}, vol. abs/1801.10130, 2018. [Online]. Available:
  \url{http://arxiv.org/abs/1801.10130}
\BIBentrySTDinterwordspacing

\bibitem{esteves2018learning}
C.~Esteves, C.~Allen-Blanchette, A.~Makadia, and K.~Daniilidis, ``Learning so
  (3) equivariant representations with spherical cnns,'' in \emph{Proceedings
  of the European Conference on Computer Vision (ECCV)}, 2018, pp. 52--68.

\bibitem{DBLP:journals/corr/abs-1812-03115}
\BIBentryALTinterwordspacing
Y.~Su and K.~Grauman, ``Kernel transformer networks for compact spherical
  convolution,'' \emph{CoRR}, vol. abs/1812.03115, 2018. [Online]. Available:
  \url{http://arxiv.org/abs/1812.03115}
\BIBentrySTDinterwordspacing

\bibitem{DBLP:journals/corr/abs-1708-00919}
\BIBentryALTinterwordspacing
------, ``Flat2sphere: Learning spherical convolution for fast features from
  360{\degree} imagery,'' \emph{CoRR}, vol. abs/1708.00919, 2017. [Online].
  Available: \url{http://arxiv.org/abs/1708.00919}
\BIBentrySTDinterwordspacing

\bibitem{zou2018layoutnet}
C.~Zou, A.~Colburn, Q.~Shan, and D.~Hoiem, ``Layoutnet: Reconstructing the 3d
  room layout from a single rgb image,'' in \emph{CVPR}, 2018.

\bibitem{yang2019dula}
S.-T. Yang, F.-E. Wang, C.-H. Peng, P.~Wonka, M.~Sun, and H.-K. Chu,
  ``Dula-net: A dual-projection network for estimating room layouts from a
  single rgb panorama,'' in \emph{Proceedings of the IEEE Conference on
  Computer Vision and Pattern Recognition}, 2019, pp. 3363--3372.

\bibitem{Sun_2019_CVPR}
C.~Sun, C.-W. Hsiao, M.~Sun, and H.-T. Chen, ``Horizonnet: Learning room layout
  with 1d representation and pano stretch data augmentation,'' in \emph{The
  IEEE Conference on Computer Vision and Pattern Recognition (CVPR)}, June
  2019.

\bibitem{Zhang_2018_ECCV}
Z.~Zhang, Y.~Xu, J.~Yu, and S.~Gao, ``Saliency detection in 360° videos,'' in
  \emph{The European Conference on Computer Vision (ECCV)}, September 2018.

\bibitem{Cheng_2018_CVPR}
H.-T. Cheng, C.-H. Chao, J.-D. Dong, H.-K. Wen, T.-L. Liu, and M.~Sun, ``Cube
  padding for weakly-supervised saliency prediction in 360° videos,'' in
  \emph{The IEEE Conference on Computer Vision and Pattern Recognition (CVPR)},
  June 2018.

\bibitem{Zioulis_2018_ECCV}
N.~Zioulis, A.~Karakottas, D.~Zarpalas, and P.~Daras, ``Omnidepth: Dense depth
  estimation for indoors spherical panoramas,'' in \emph{The European
  Conference on Computer Vision (ECCV)}, September 2018.

\bibitem{DBLP:journals/corr/abs-1811-05304}
\BIBentryALTinterwordspacing
F.~Wang, H.~Hu, H.~Cheng, J.~Lin, S.~Yang, M.~Shih, H.~Chu, and M.~Sun,
  ``Self-supervised learning of depth and camera motion from 360{\degree}
  videos,'' \emph{CoRR}, vol. abs/1811.05304, 2018. [Online]. Available:
  \url{http://arxiv.org/abs/1811.05304}
\BIBentrySTDinterwordspacing

\bibitem{Garanderie_2018_ECCV}
G.~Payen~de La~Garanderie, A.~Atapour~Abarghouei, and T.~P. Breckon,
  ``Eliminating the blind spot: Adapting 3d object detection and monocular
  depth estimation to 360° panoramic imagery,'' in \emph{The European
  Conference on Computer Vision (ECCV)}, September 2018.

\bibitem{sweepnet}
C.~{Won}, J.~{Ryu}, and J.~{Lim}, ``Sweepnet: Wide-baseline omnidirectional
  depth estimation,'' in \emph{2019 International Conference on Robotics and
  Automation (ICRA)}, May 2019, pp. 6073--6079.

\bibitem{Eder_2019}
\BIBentryALTinterwordspacing
M.~Eder, P.~Moulon, and L.~Guan, ``Pano popups: Indoor 3d reconstruction with a
  plane-aware network,'' \emph{2019 International Conference on 3D Vision
  (3DV)}, Sep 2019. [Online]. Available:
  \url{http://dx.doi.org/10.1109/3DV.2019.00018}
\BIBentrySTDinterwordspacing

\bibitem{Zioulis_2019}
\BIBentryALTinterwordspacing
N.~Zioulis, A.~Karakottas, D.~Zarpalas, F.~Alvarez, and P.~Daras, ``Spherical
  view synthesis for self-supervised 360° depth estimation,'' \emph{2019
  International Conference on 3D Vision (3DV)}, Sep 2019. [Online]. Available:
  \url{http://dx.doi.org/10.1109/3DV.2019.00081}
\BIBentrySTDinterwordspacing

\bibitem{chen2014semantic}
L.-C. Chen, G.~Papandreou, I.~Kokkinos, K.~Murphy, and A.~L. Yuille, ``Semantic
  image segmentation with deep convolutional nets and fully connected crfs,''
  \emph{arXiv preprint arXiv:1412.7062}, 2014.

\bibitem{savva2017minos}
M.~Savva, A.~X. Chang, A.~Dosovitskiy, T.~Funkhouser, and V.~Koltun, ``{MINOS}:
  Multimodal indoor simulator for navigation in complex environments,''
  \emph{arXiv:1712.03931}, 2017.

\bibitem{wang2018self}
F.-E. Wang, H.-N. Hu, H.-T. Cheng, J.-T. Lin, S.-T. Yang, M.-L. Shih, H.-K.
  Chu, and M.~Sun, ``Self-supervised learning of depth and camera motion from
  360\degree videos,'' in \emph{Asian Conference on Computer Vision}.\hskip 1em
  plus 0.5em minus 0.4em\relax Springer, 2018, pp. 53--68.

\bibitem{kannala2006generic}
J.~Kannala and S.~S. Brandt, ``A generic camera model and calibration method
  for conventional, wide-angle, and fish-eye lenses,'' \emph{IEEE transactions
  on pattern analysis and machine intelligence}, vol.~28, no.~8, pp.
  1335--1340, 2006.

\end{thebibliography}
